\documentclass{article}
\pdfoutput=1
\usepackage[nonatbib,final]{nips_2016}

\usepackage[utf8]{inputenc} 
\usepackage[T1]{fontenc}    
\usepackage{url}            
\usepackage{booktabs}       
\usepackage{amsfonts}       
\usepackage{nicefrac}       
\usepackage{microtype}      
\usepackage{algorithm}
\usepackage{algorithmic}
\usepackage{amsmath}
\usepackage{bm}
\usepackage[usenames, dvipsnames]{color}
\usepackage[pdftex]{graphicx}
\usepackage{subfigure} 
\title{Automatic Variational ABC}

\usepackage{authblk}

\author[1]{\textbf{Alexander Moreno}\thanks{corresponding author, alexander.f.moreno@gmail.com} }
\author[2,3]{\textbf{Tameem Adel}}
\author[2,4]{\textbf{Edward Meeds}}
\author[1]{\textbf{James M. Rehg}}
\author[2,5]{\textbf{Max Welling}}
\affil[1]{College of Computing, Georgia Institute of Technology}
\affil[2]{Machine Learning Group, University of Amsterdam}
\affil[3]{Data Science, Radboud University}
\affil[4]{Center for Integrative Bioinformatics IBIVU, VU University}
\affil[5]{Canadian Institute for Advanced Research (CIFAR)}

\begin{document}

\maketitle

\begin{abstract}
Approximate Bayesian Computation (ABC) is a framework for performing likelihood-free posterior inference for simulation models. Stochastic Variational inference (SVI) is an appealing alternative to the inefficient sampling approaches commonly used in ABC. However, SVI is highly sensitive to the variance of the gradient estimators, and this problem is exacerbated by approximating the likelihood.  We draw upon recent advances in variance reduction for SVI \cite{kingma2014auto}\cite{ranganath2013black} and likelihood-free inference using deterministic simulations \cite{meeds2015optimization} to produce low variance gradient estimators of the variational lower-bound.  By then exploiting automatic differentiation libraries \cite{autograd} we can avoid nearly all model-specific derivations. We demonstrate performance on three problems and compare to existing SVI algorithms.  Our results demonstrate the correctness and efficiency of our algorithm.
\end{abstract}

\section{Introduction}\label{introduction}

In many areas of science complex simulators are used as models of the underlying phenomena of interest.  For example, in computational biology, models could be simulations of embryonic morphogenesis or cancer development; in environmental science, they could model earthquakes or climate change.  Further examples of simulation-based science can be found in computational chemistry, physics, and neuroscience.  
The most fundamental ingredient and computational bottleneck is the ability to match a (simulation) model to given observations. Bayesian methods provide an elegant solution to this problem where posterior distributions provide insightful information about the correlations and sensitivities of the parameters. However, current methods are based on sampling, such as MCMC, which tends to converge slowly and requires many calls to the simulator. In this work we introduce a variational Bayesian alternative. 

%
%
%
%
In Bayesian inference, one wants to infer the posterior distribution $p(\bm{\theta}|\bf{y})$ of some parameters $\bm{\theta}$, given observations $\bf{y}$.
%
%
%
\begin{align}
p(\bm{\theta}|\bm{y})&=\frac{p(\bm{y}|\bm{\theta})p(\bm{\theta})}{\int p(\bm{y}|\bm{\theta})p(\bm{\theta})d\bm{\theta}}
\end{align}
%
%
%
where $p(\bm{y}|\bm{\theta})$ is the likelihood, $p(\bm{\theta})$ is a prior, and $\int p(\bm{y}|\bm{\theta})p(\bm{\theta})d\bm{\theta}$ is the normalizing constant. The normalizing constant is often intractable, so that sampling methods or variational inference should be used. However, both assume a tractable expression for the likelihood. For many problems, the likelihood is intractable or extremely expensive to compute.  Approximate Bayesian Computation (ABC) deals with how to do Bayesian inference by approximating the likelihood using simulator outputs. The simulator generates synthetic data $\bm{x}$ according to the parameters $\bm{\theta}$, i.e. we can simulate $\bm{x}\sim p(\bm{x}|\bm{\theta})$.  These are compared to the true data via an $\epsilon$-kernel or ABC-kernel $K_{\epsilon}(y,x)$, which is an approximation to $p(\bm{y}|\bm{x})$ and is a density that measures the discrepancy between $\bm{x}$ and $\bm{y}$, where $\epsilon$ is the bandwidth. Often, $K_{\epsilon}(\bm{y},\bm{x})= \mathcal{N}(\bm{y}|\bm{x},\epsilon^{2})$, but one may take other distributions.  The likelihood, $p(\bm{y}|\bm{\theta})$, is approximated by the ABC likelihood
%
%
%
\begin{align}
p_{\epsilon}(\bm{y}|\bm{\theta})&=\int K_{\epsilon}(\bm{y},\bm{x}) p( \bm{x}|\bm{\theta}) \: dx\\
&\approx \frac{1}{S}\sum_{s=1}^{S}K_{\epsilon}(\bm{y},\bm{x}^{(s)})
\end{align}
The bias of $p_{\epsilon}(\bm{y}|\bm{\theta})$ goes to $0$ as $\epsilon\rightarrow 0$.  With large and/or high-dimensional observations, it may be beneficial to use summary statistics $S(\bm{y})$ and $S(\bm{x})$ instead of the entire data set, i.e. $K_{\epsilon}(S(\bm{y}),S(\bm{x}))$ instead of $K_{\epsilon}(\bm{y},\bm{x})$.  $\epsilon$ introduces a trade-off between bias with a large $\epsilon$ and variance with a small $\epsilon$.  We will assume that $\bm{y}$ and $\bm{x}$ will, depending on the context,  represent either raw data or statistics.

Sampling methods are widely used in the likelihood-free and ABC literature: rejection sampling \cite{tavare1997inferring}, Markov Chain Monte Carlo (MCMC) via a modified Metropolis Hastings algorithm \cite{marjoram2003markov}, and population-based sampling \cite{beaumont2009adaptive,del2006sequential,sisson2007sequential}.  However, these have slow mixing rates and make many calls to the simulator, making them ineffective for large-scale problems, although some recent methods have been proposed to improve this, including \cite{Meeds2014GpsUai}\cite{wilkinson:2014}.  Stochastic variational inference (SVI) typically has a faster rate of convergence but might still suffer from a high variance in the estimated gradients, as we will see.

\section{Related Work}

\cite{kingma2014auto} \cite{rezende2014stochastic} showed that for certain classes of continuous latent variables and variational posteriors, by reparametrizing the parameters or latent variables to be estimated, one can obtain a lower variance Monte Carlo approximation to the gradient of the lower bound.  \cite{titsias2015local} found a more general variance reduction method that is conceptually similar to Gibbs sampling, but tends to perform worse when using the same variational posterior.

%
%
\cite{ranganath2013black} (BBVI) showed how to do stochastic variational inference in the non-conjugate case for both discrete and continuous latent variables with minimal model-specific derivation.  This involves taking a Monte-Carlo approximation of the gradient of the expected lower bound, using ADAGRAD to avoid taking derivatives by hand, and using Rao Blackwellization with a mean-field assumption and control variates to reduce the variance of the gradients.  \cite{ruiz2016overdispersed} (O-BBVI) extended this work by adding importance sampling to further reduce the variance of the gradient estimator.  \cite{tran2015variational} (VBIL) builds on the same naive estimator as BBVI, but uses natural gradient descent instead of ADAGRAD and treats the ABC case.  They show that despite a bias being introduced by taking the log of a likelihood estimator, it is equivalent to using the true likelihood, and they apply this to several intractable likelihood problems.  See the supplementary material for the similarity between BBVI and VBIL.

\cite{kucukelbir2015automatic} used automatic differentiation to fully automate variational inference, which we refer to as AVI.  AVI fits a Gaussian variational distribution in a transformed space, leading to a non-Gaussian approximation in the original space when transformed back.  This limits the range of variational distributions.  In the transformed space they make a fully-factorized mean-field assumption.  AVI requires rewriting the simulator in STAN.  With complex simulators, this is unrealistic and negates the main benefit: that it is automatic.

\section{Automatic Variational ABC}

\subsection{The Variational Lower Bound}\label{sec:vlb}

In this section we will derive the variational lower bound for our ABC problem and use a number of methods to reduce the variance in estimating its gradients. 

Consider a simulator model $p(\bm{x}|\bm{\theta})$ for which we do not have an analytic expression available but which is given to us as a (potentially complex) piece of computer code (otherwise known as a ``probabilistic program''). Here, $\bm{x}$ is the output of our simulator, or some summary statistics thereof (the computation of which we will assume to be a part of the simulator) and $\bm{\theta}$ are parameters of the model. The data's marginal likelihood is given by
\begin{align}
\log p(\bm{y})&=\log\int p(\bm{y}|\bm{\theta})p(\bm{\theta})d\bm{\theta}\\
&=\log\int Q_\phi(\bm{\theta})\frac{p(\bm{y}|\bm{\theta})p(\bm{\theta})}{Q_\phi(\bm{\theta})}d\bm{\theta}\\
&\geq \int Q_\phi(\bm{\theta})\log \frac{p(\bm{y}|\bm{\theta})p(\bm{\theta})}{Q_\phi(\bm{\theta})}d\theta\textrm{ by Jensen's inequality}
\end{align}

this assumes a variational posterior $Q_\phi(\bm{\theta})$ over the variable $\theta$.  We denote the last term as $\mathcal{L}$ to indicate a lower bound on the marginal probability.  Maximizing $\mathcal{L}$ is equivalent to minimizing the KL divergence between the variational posterior and the true posterior $KL(Q_\phi(\bm{\theta})||p(\bm{\theta}|\bm{y}))$.  We then have
\begin{align}
\mathcal{L}&=\int Q_\phi(\bm{\theta})\log \frac{p(\bm{y}|\bm{\theta})p(\bm{\theta})}{Q_\phi(\bm{\theta})}d\bm{\theta}\\
&=\int Q_\phi(\bm{\theta})\log p(\bm{y}|\bm{\theta})d\bm{\theta}-KL(Q_\phi(\bm{\theta})||p(\bm{\theta}))\\
&=\mathbb{E}_Q(\log p(\bm{y}|\bm{\theta}))-KL(Q_\phi(\bm{\theta})||p(\bm{\theta}))\label{eq:avabc}
\end{align}

We now replace the true likelihood with the ABC likelihood, giving
\begin{align}
\mathcal{L_{ABC}}&=\int Q_\phi(\bm{\theta})\log \int p_\epsilon(\bm{y}|\bm{x})p(\bm{x}|\bm{\theta})d\bm{x}d\bm{\theta}-KL(Q_\phi(\bm{\theta})||p(\bm{\theta}))\\
&=\int Q_\phi(\bm{\theta})\log \int p_\epsilon(\bm{y}|f(\bm{\theta},\bm{u}))p(\bm{u})d\bm{u}d\bm{\theta}-KL(Q_\phi(\bm{\theta})||p(\bm{\theta}))
\end{align}

where $p_\epsilon(\bm{y}|f(\bm{\theta},\bm{u}))$ is the $\epsilon$-kernel and we have used our first reparameterization \cite{Meeds2015HABC,meeds2015optimization} where we replace 
\begin{align}
& \int d\bm{x} ~p(\bm{x}|\theta) H[\bm{x}]  = \int d\bm{u} ~ p(\bm{u}) H\left[f(\bm{\theta},\bm{u})\right]
\end{align}
with $H$ some arbitrary function of $\bm{x}$. The simulator $p(\bm{x}|\bm{\theta})$ is now written as a deterministic function $f(\bm{\theta},\bm{u})$ where the randomness is now externalized into a variable $\bm{u}\sim p(\bm{u})$. For example, if $p(\bm{x}|\bm{\theta})$ is normally distributed and $\bm{\theta}$ contains the mean and variance, then $\bm{x}=\bm{\mu}+ \bm{\Sigma} \bm{u}$ with $\bm{u}\sim N(\bm{0},\bm{I})$.  Once again we will assume that the simulator output is either the raw data or a vector of statistics; in both cases we will represent the deterministic version as $\bm{x} = f(\bm{\theta},\bm{u})$.
To reduce the variance of the gradients, we apply a second reparametrization \cite{kingma2014auto}:
\begin{align}\label{eq:vab1n}
\int d\bm{\theta} ~Q_\phi(\bm{\theta}) H[\bm{\theta}] &= \int d\bm{\nu} ~ Q_0(\bm{\nu}) H\left[g(\bm{\phi},\bm{\nu})\right]\\
\end{align}
where it is important to note that the distribution $Q_0$ is constant w.r.t. any parameters of interest, and all dependency on the parameters $\bm{\phi}$ is transferred to the function $g$. This gives
\begin{align}
\mathcal{L_{ABC}}&=\int Q_0(\bm{\nu})\log \int p_\epsilon(\bm{y}|f(g(\bm{\phi},\bm{\nu}),\bm{u}))p(\bm{u})d\bm{u}d\bm{\nu}-KL(Q_\phi(\bm{\theta})||p(\bm{\theta}))
\end{align}
Assuming that $KL(q(\bm{\theta})||p(\bm{\theta}))$ can be calculated analytically, we replace expectations by samples to obtain
\begin{align}
\mathcal{L_{ABC}}&\approx\frac{1}{S}\sum_{s=1}^{S}\log \frac{1}{L}\sum_{l=1}^{L} p_\epsilon(\bm{y}|f(g(\bm{\phi},\bm{\nu}^{(s)}),\bm{u}^{(l)}))-KL(Q_\phi(\bm{\theta})||p(\bm{\theta}))\label{eq:lowerbound}
\end{align}

where $\bm{\nu}^{(s)}\sim Q_0(\bm{\nu})$ and $\bm{u}^{(l)}\sim p(\bm{u})$.

When $KL(Q_\phi(\bm{\theta})||p(\bm{\theta}))$ cannot be calculated analytically, we can apply the second reparametrization to it and approximate it via sampling.

One issue is that even assuming that the ABC likelihood is an unbiased estimator of the true likelihood (which it is not), taking the log introduces a bias, so that we now have a biased estimate of the lower bound and thus biased gradients.  We show in the appendix that the theory developed in VBIL applies to our estimator, and thus taking the log of an estimator is not a problem.  In the appendix we extend this model to learning a variational distribution for latent variables per datapoint.

\subsection{Automatic Differentiation and Averaged Gradients}

We now wish to take derivatives w.r.t. $\bm{\phi}$ to compute stochastic gradients and ascent on the bound eq.~\ref{eq:lowerbound}.  This requires differentiating through $\log p_{\epsilon}(\bm{y}| f(g(\bm{\phi},\bm{\nu}^{(s)}),\bm{u}) )$, and via the chain rule, also through the simulator function $f(\bm{\theta},\bm{u})$.  Because we write our simulator code within an auto-differentiation environment (in our case, Python, where we can apply Autograd~\cite{autograd}), this is handled automatically by the optimization algorithm.  This avoids the model specific derivation of taking derivatives by hand, while allowing us to use a wider range of variational posteriors than \cite{kucukelbir2015automatic}.  A strong requirement that will be discussed further in the experiments, is that the function $g$ can deterministically generate $\bm{\theta}$ and $\bm{u}$, which may not be possible for all distributions of interest.  


In our experiments we use adaptive gradient algorithms which automatically tune the learning rates per parameter according to the history of gradients and their variances.  Two such algorithms are ADAM \cite{kingma2014adam} and ADAGRAD \cite{duchi2011adaptive}: we used the latter in our experiments.  As we will discuss later, using these algorithms greatly reduces the effect of large variance gradients and is especially helpful for VBIL experiments.  Algorithm \ref{alg:alfvi} describes the procedure for sampling from parameterless distributions, differentiating the lower bound, and applying an adaptive gradient-step.

Another important challenge is choosing $\epsilon$.  As we increase $\epsilon$, the bias of the likelihood and thus our gradient estimator goes up, but as we decrease it, the variance goes up.  Wilkinson~\cite{Wilkinson2013} showed that in the likelihood-free setting, $\epsilon$ could be interpreted as model or simulation noise.  We can empirically take advantage of this interpretation and set $\epsilon$ according to the standard deviation of the mean of the raw simulation data, depending upon whether this is appropriate for the statistics used.  For example, if $\bm{x}$ is the length $M$ vector of raw data from the simulator we can set $\epsilon =\sigma(\bm{x})/\sqrt[]{M}$.  


%
\begin{algorithm}[tb]
   \caption{Automatic Variational ABC}
   \label{alg:alfvi}
\begin{algorithmic}
   \STATE Specify simulation code $f(\bm{\theta},\bm{u})$ and simulator randomness $p(u)$
   \STATE Specify reparametrization $Q(\bm{\theta}|\bm{\bm{\phi}})$ in terms of parameterless $Q_{0}(\bm{\nu})$
   \STATE $\bm{\bm{\phi}}\leftarrow$ initialize parameters
   \REPEAT
   \STATE $\bm{\nu}\leftarrow$ samples from $Q_{0}(\bm{\nu})$
   \STATE $\bm{u}\leftarrow$ samples from $p(u)$
   \STATE $g\leftarrow\nabla \tilde{\mathcal{L}}_{\bm{\bm{\phi}}}(\bm{\bm{\phi}},\bm{\nu},u)$
   \STATE $\bm{\phi}\leftarrow$ run one step of an adaptive gradient algorithm using $g$.
   \UNTIL{convergence}
   \STATE return $\bm{\bm{\phi}}$
\end{algorithmic}
\end{algorithm}

\section{Experiments}

We perform three experiments: 1) inferring the success probability from Bernoulli trials 2) inferring the rate of an exponential distribution 3) inferring the parameters of a stochastic biological system of blowfly populations.  For each experiment, we use a Gaussian kernel for the ABC-likelihood, and adaptively set $\epsilon$ depending on the simulation problem.  We compare to BBVI, which as mentioned before, is the same as VBIL but using ADAGRAD \cite{duchi2011adaptive} instead of natural gradient descent.  We use ADAM for both our method and BBVI (in the original paper, they used ADAGRAD).  Because of the posterior limitations and requirement of rewriting a potentially complex simulator in STAN, we do not compare to AVI.  Due to higher variance, running BBVI occasionally leads to invalid parameter values; when that happens, we reinitialize.

\subsection{Bernoulli Problem}

We perform the first VBIL experiment.  We have $M$ Bernoulli samples, each with success probability $\bm{\theta}$, and we infer a distribution for $\bm{\theta}$.  Our observed data $\bm{y}$, is a vector of $1$'s and $0$'s.  Let $k=\sum_{m=1}^{M}y_{m}$. With a Beta$(1,1)$ prior, the true posterior is Beta$(k+1,M-k+1)$.  Because we need to differentiate with respect to the simulator, instead of simulating $M$ Bernoulli trials, which gives discrete outputs, we use the normal approximation to the binomial with $M$ trials and success probability $\bm{\theta}$, and sample from that.  Thus $S(\bm{x})$ is simply the output of the simulator.  We set $M=100$ and $k=70$.  For this problem, we cannot calculate the sample variance as in the $\epsilon$-selection method described.  However, we note that under the Gaussian approximation to the binomial, the simulator variance is $M\cdot\theta(1-\theta)$, so we set $\epsilon=\sqrt[]{\theta(1-\theta)}$.

For both AVABC and BBVI, we use a Kumaraswamy distribution as the variational posterior for $\bm{\theta}$, since the Kumaraswamy distribution is similar to the Beta, but has a closed form inverse CDF, allowing us to generate samples deterministically given samples from the uniform distribution.  Thus, $Q_{0}(\bm{\nu}) = \text{U}(0,1)$.  For the Gaussian approximation in the simulation, we use $\mathcal{N}(\bm{0},\bm{I})$.  We initialize by setting the Kumaraswamy parameters to be $(1,1)$.
%

\begin{figure}
\centering
\subfigure[lower bounds]{
\includegraphics[width=0.4\textwidth,height=0.3\textwidth]{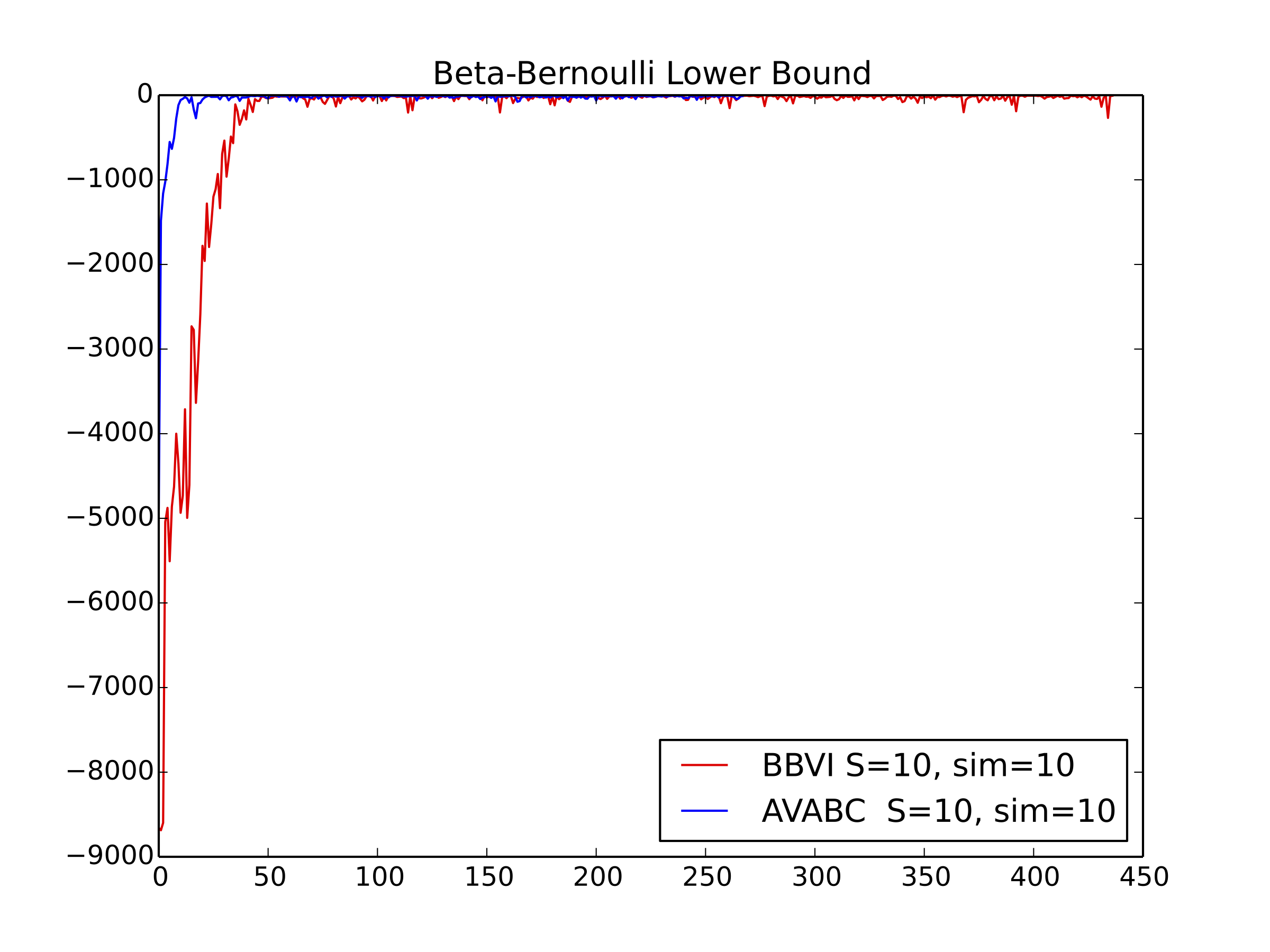}\label{fig:beta01}}
\quad
\subfigure[posteriors]{
\includegraphics[width=0.4\textwidth,height=0.3\textwidth]{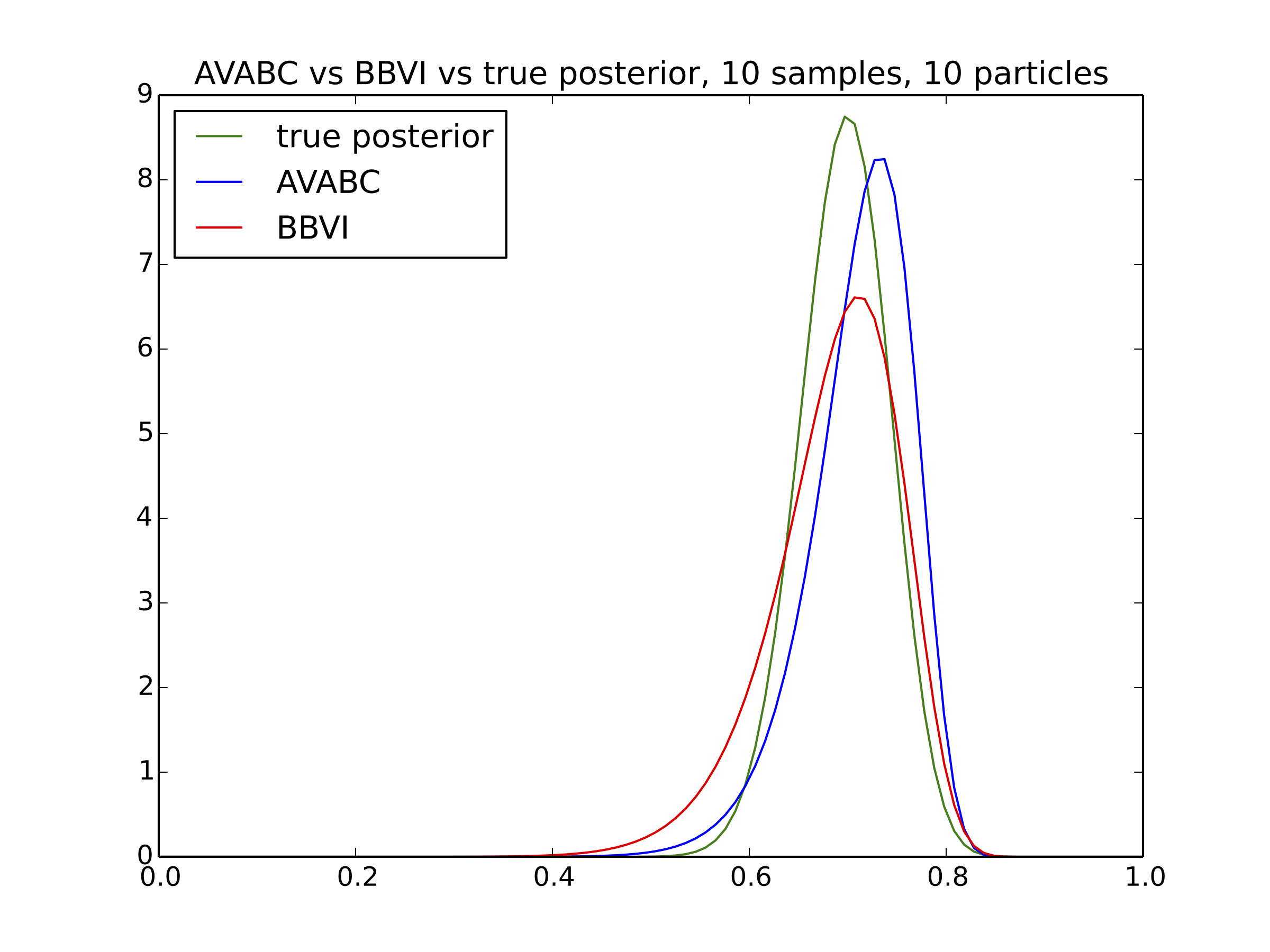}\label{fig:beta02}}
\quad
\subfigure[gradient distribution, convergence parameters]{
\includegraphics[width=0.4\textwidth,height=0.3\textwidth]{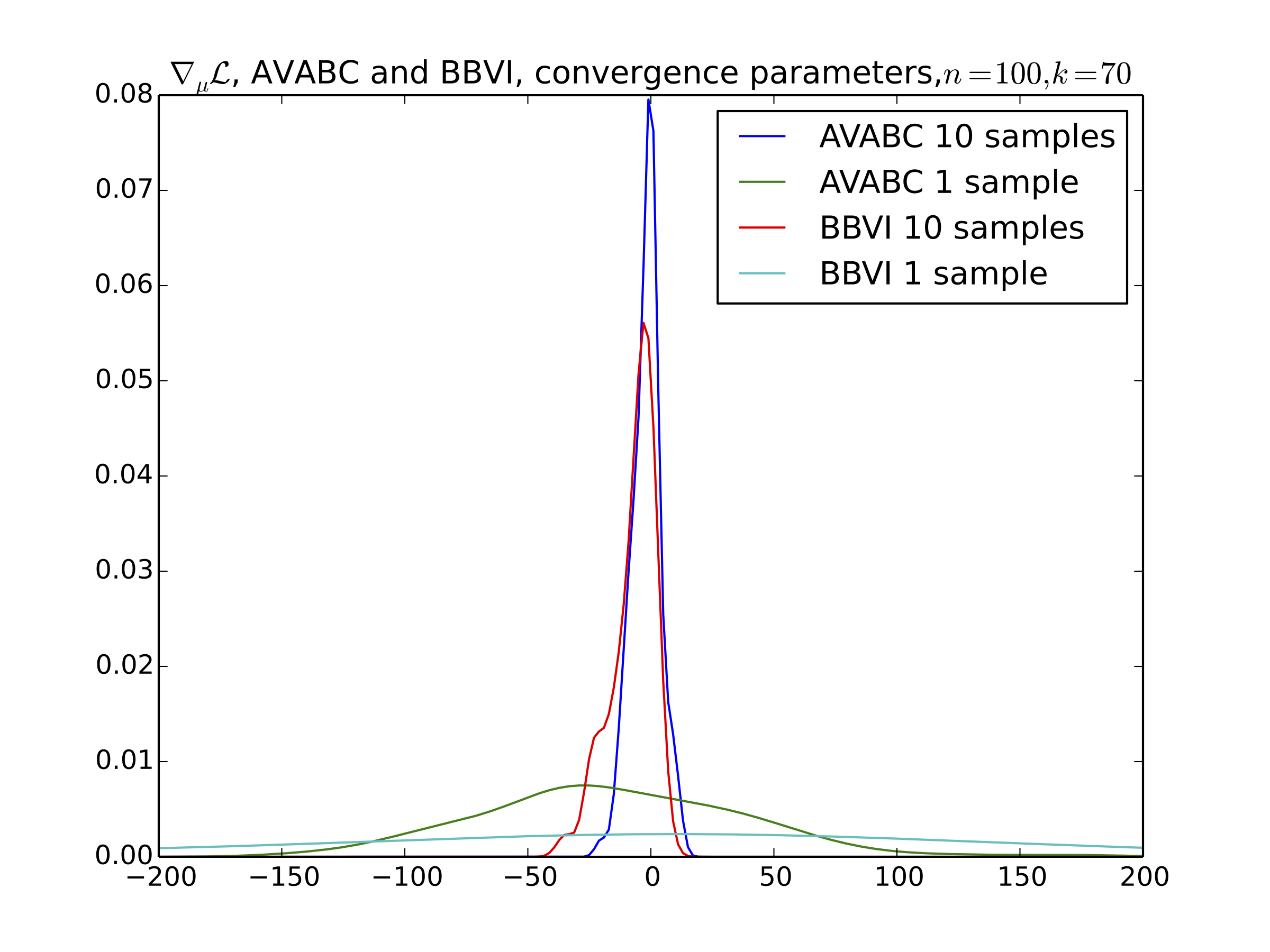}\label{fig:beta03}}
\quad
\subfigure[gradient distribution for an algorithm run, from start until convergence]{
\includegraphics[width=0.4\textwidth,height=0.3\textwidth]{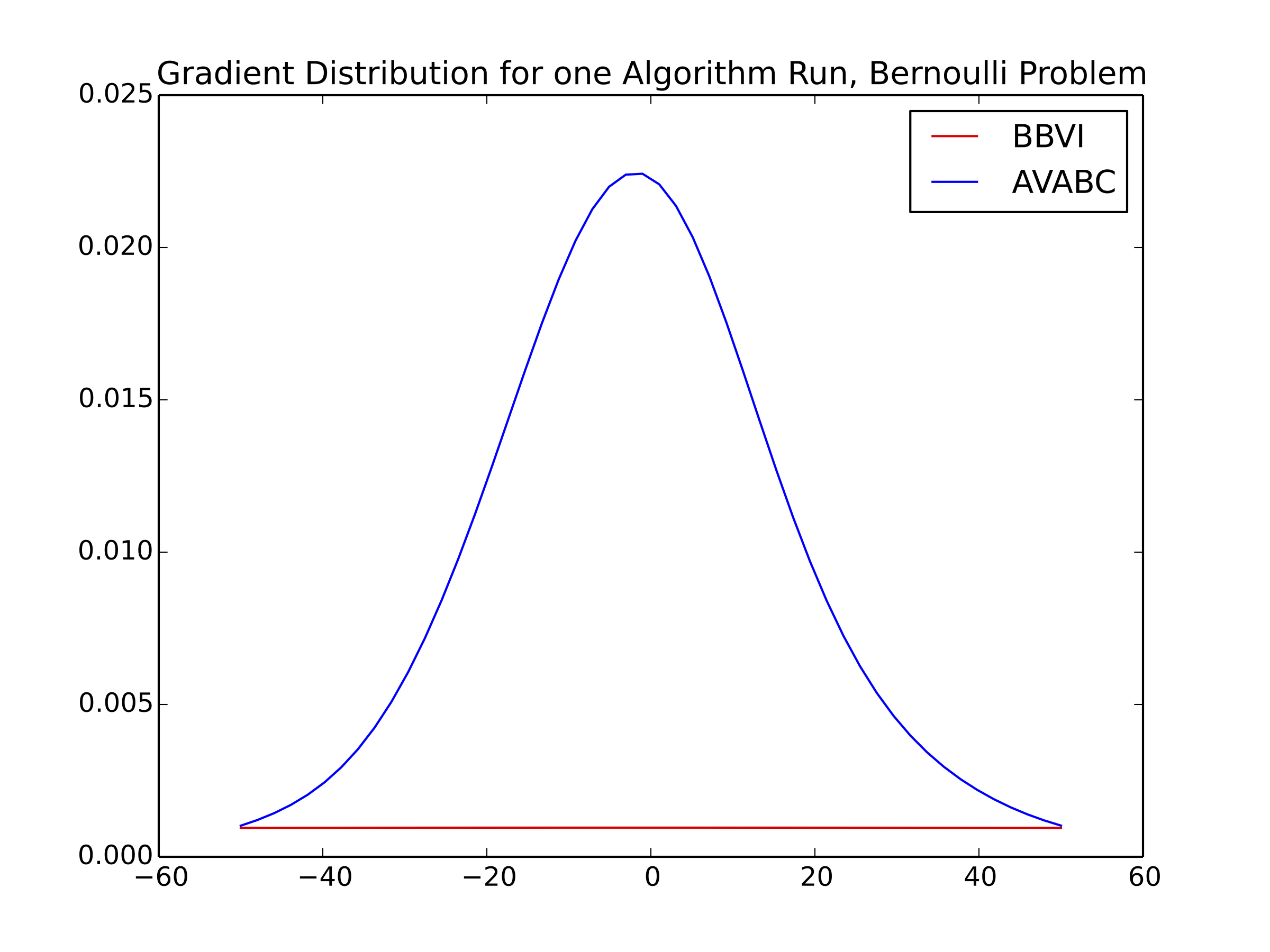}\label{fig:beta04}}
\label{fig:betafig}
\caption{a) Lower bounds for the Bernoulli problem.  Performance is similar between the two methods.  Both methods stabilize in about 100 iterations.  b) posteriors for both methods compared to the true posterior.  Both achieve decent posteriors, despite using a different variational distribution from the true posterior family. c) Gradients for the Bernoulli problem for $\mu$ (chosen because it has higher variance gradients than $\sigma$) at convergence values.  Our model with 10 samples is in blue, while for BBVI it is in red.  For 10 samples, they have similar gradient variance.  However, our model with 1 sample is in green, while BBVI with 1 sample is turqoise, showing that for a single sample our model has far lower variance. d) gradient distribution for the Bernoulli problem for $\mu$ across a full run of the algorithm.}
\end{figure}

Figure \ref{fig:beta01} shows the lower bound for the Beta problem.  Both methods have their lower bounds stabilize in under 100 iterations.  \ref{fig:beta02} shows the posteriors: both are reasonable.  \ref{fig:beta03} shows the naive gradient distribution taking 100 samples at the convergence parameters.  We see that for $10$ samples and simulations per sample, the gradients are approximately equal, but for $1$ sample and simulation, AVABC has far lower variance than BBVI.  \ref{fig:beta04} shows the naive gradient distributions for a full run of the algorithm.  That is, we run each algorithm from start until convergence, store the gradients, and plot the distributions.  AVABC has far lower variance.  ADAM drastically reduces the variance of both methods so that they are almost equivalent, leading to the similar convergence.
\subsection{Exponential Problem}

We wish to infer a distribution for the rate $\lambda$ of an exponential distribution.  In this setting, with a Gamma$(\alpha,\beta)$ prior and a data vector $y$ of $M$ samples from the exponential distribution, the true posterior is Gamma$(\alpha+M,\beta+\bar{y}M)$.  The simulator draws samples from the exponential distribution.  We use statistics $S(\bm{x})=\bar{x}$ and $S(\bm{y})=\bar{y}$.  We set the true $\lambda=1$, and $M=15$.

For both methods, we use a log-normal variational posterior model.  At each iteration, we can take samples $\bm{\nu}^{(s)}$ from $Q_{0}(\bm{\nu})\sim \mathcal{N}(0,I)$ and then deterministically calculate
\begin{align}
\theta = g(\bm{\phi},\bm{\nu}^{(s)})=\exp (\mu+\Sigma\cdot\bm{\nu_{s}})
\end{align}
We initialize both $\mu$ and $\sigma$ by drawing from $\text{U}(2,3)$.  The convergence parameters are approximately $\mu=0$ and $\sigma=0.3$, so given that we use the lognormal distribution, these parameters are fairly far away.  We use $10$ samples and $10$ particles/simulations per sample.  That is, $S=10$ and $L=10$.

\begin{figure}[ht]
\centering
\subfigure[lower bounds]{
\includegraphics[width=0.4\textwidth,height=0.3\textwidth]{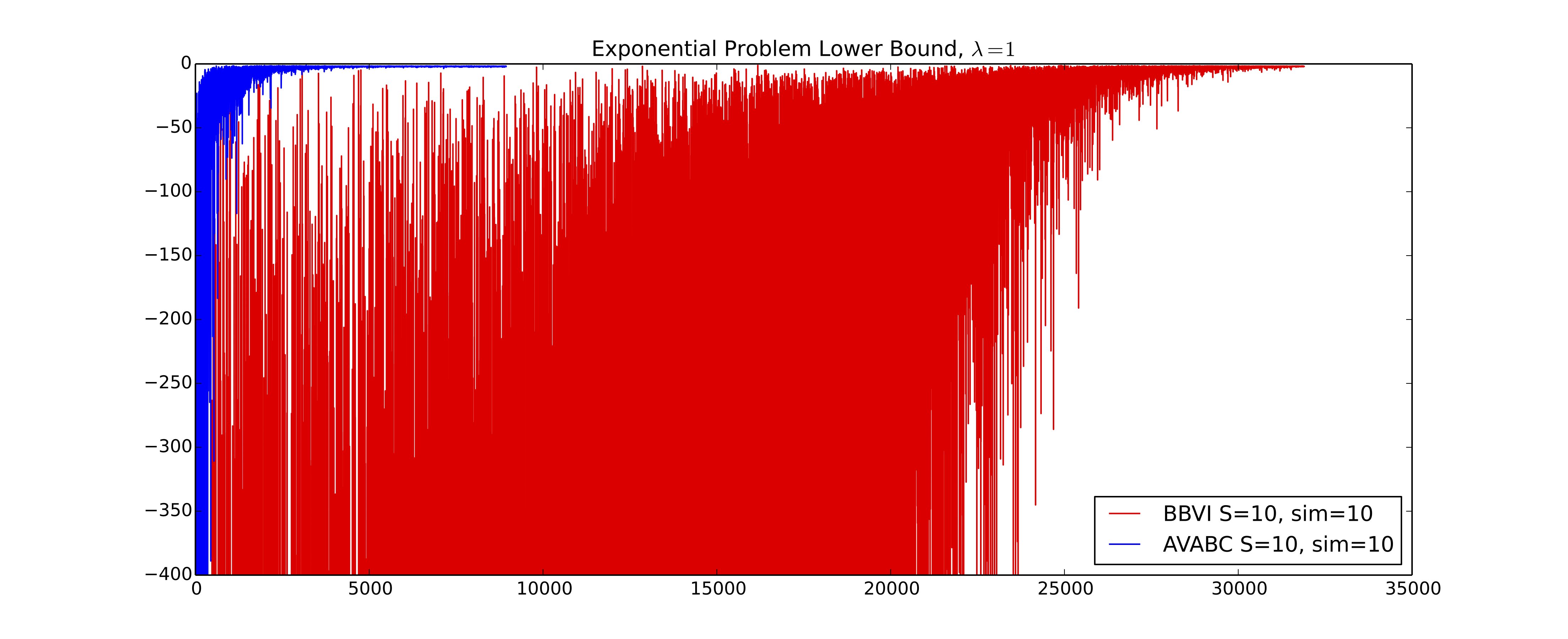}\label{fig:ef01}}
\quad
\subfigure[posteriors]{
\includegraphics[width=0.4\textwidth,height=0.3\textwidth]{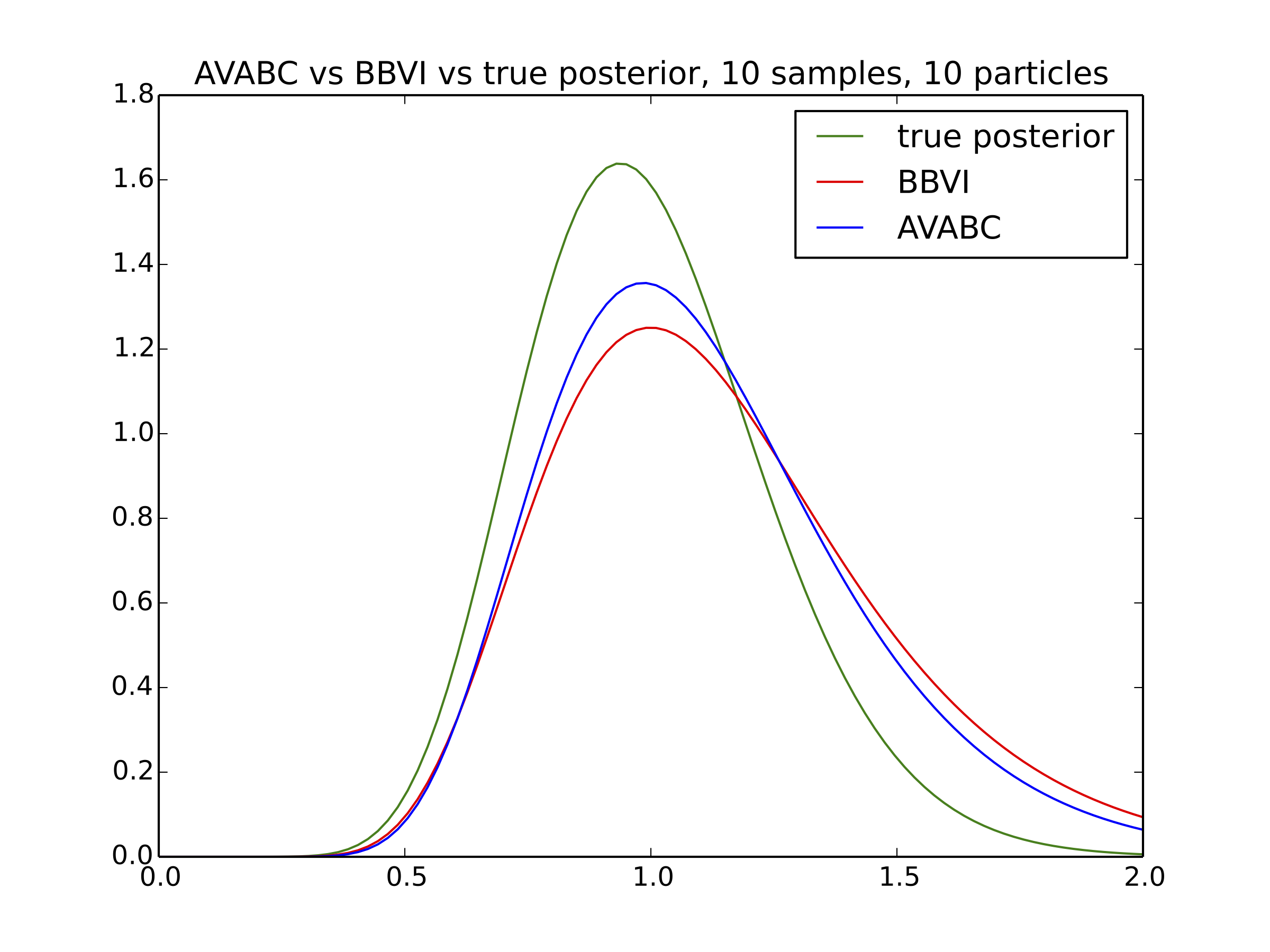}\label{fig:ef02}}
\quad
\subfigure[gradient distribution]{
\includegraphics[width=0.4\textwidth,height=0.3\textwidth]{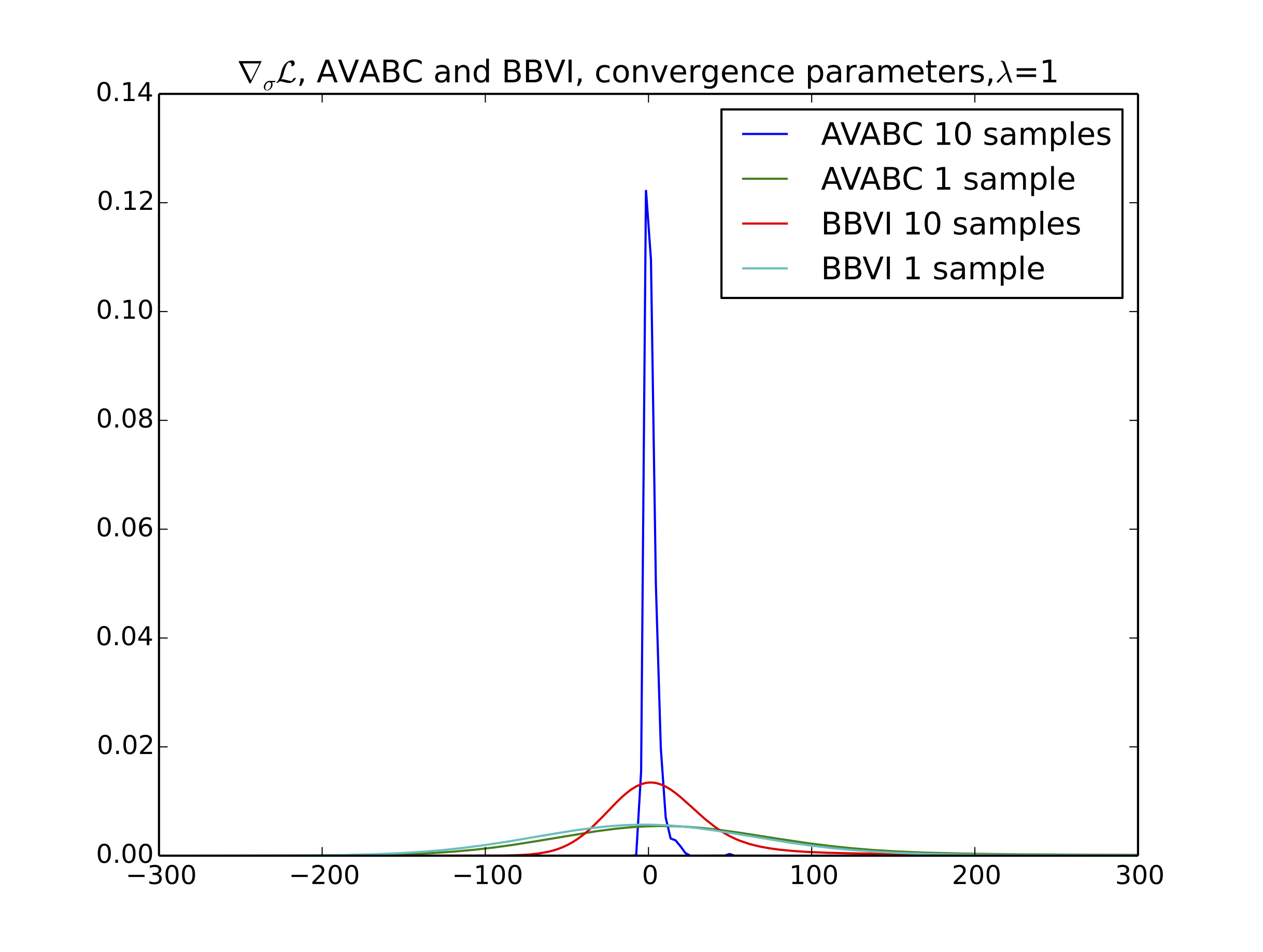}\label{fig:ef03}}
\caption{a) Lower bounds for the exponential problem with 10 samples and 10 particles/simulations per sample.  BBVI (red) has drastically higher variance and takes substantially more iterations to converge than AVABC (blue).  b) posteriors for both methods compared to the true posterior c) Gradients for the exponential problem at convergence values.  Our model with 10 samples is in blue and is the most peaked, showing that with 10 samples, our model has far lower variance close to a local optimum.}
\end{figure}

Figures \ref{fig:ef01} shows the lower bound plot for the exponential problem.  AVABC, in blue, has much lower variance and faster convergence.  AVABC stabilizes at around $4,000$ iterations, while BBVI takes over $30,000$ iterations to stabilize.  \ref{fig:ef02} shows the posteriors against the true posterior (green).  Both have relatively good estimates.  \ref{fig:ef03} shows the distribution of gradients for both a single sample and 10 samples.  For $10$ samples, the AVABC gradient distribution has far lower variance and is thus more peaked.


\subsection{Blowfly Problem}
Performing well-founded statistical inference in biological dynamic models representing chaotic and near-chaotic systems is difficult. We apply our method to a problem where the dynamic behavior of adult blowfly populations is analyzed. We use the observed data of \cite{Wood2010} where discretized differential equations are used to model the blowfly population dynamics. Parameter settings resulting from this modeling can have some chaotic behavior. We base our simulation on the following -equation (1) in Section 1.2.3 of the supplementary material in \cite{Wood2010}-:
\begin{align}\label{eq:bfly01}
N_{t+1} = P N_{t - \tau} \exp(\frac{-N_{t - \tau}}{N_0}) e_t + N_t \exp(- \delta \epsilon_t)
\end{align}
where $e_t \! \sim \! \mathcal{G}(1/\sigma_p^2, 1/\sigma_p^2)$ and $e_t \sim \mathcal{G}(1/\sigma_d^2, 1/\sigma_d^2)$ are noise sources, and $\tau$ is an integer. There are $5$ parameters to estimate; $\bm{\theta} = \{\log P, \log \delta, \log N_0, \log \sigma_d, \log \sigma_p \}$. Similar to \cite{Meeds2014GpsUai,Meeds2015HABC}, priors of $\bm{\theta}_{1 \cdots 5}$ are Gaussian distributions. The vector of observations, $\bm{y}$, consists of a time-series of daily counts of a blowfly population. Generating data $\bm{x}$, from a simulator based on parameters, $\bm{\theta}$, of a time-series, is an intriguing task, since, in addition to their chaotic nature, small changes in the parameter prior(s) might lead to degenerate $\bm{x}$. There are $10$ statistics in use (again similar to \cite{Meeds2014GpsUai}): the mean values of each of the $4$ quartiles of the simulated samples, the mean values of the $4$ quartiles of the first-order differences of the simulated samples and the number of maximal peaks under $2$ different thresholds.

Our prior and Q distribution are Gaussian:
\begin{align}\label{eq:bfly02}
p(\bm{\theta}) &= \mathcal{N}( \bm{\mu}, \bm{\sigma}) \notag \\
Q(\bm{\theta}|\bm{\phi}) &= \mathcal{N}(\bm{\mu}_{\bm{\phi}}, \bm{\sigma}_{\bm{\phi}})
\end{align}
where $\mathcal{N}$ refers to a Gaussian distribution and therefore $g$ is  the transformation of standard normals.

\begin{figure}[ht]
\centering
\subfigure[$\log P$]{%
\includegraphics[width=37mm,height=41mm]{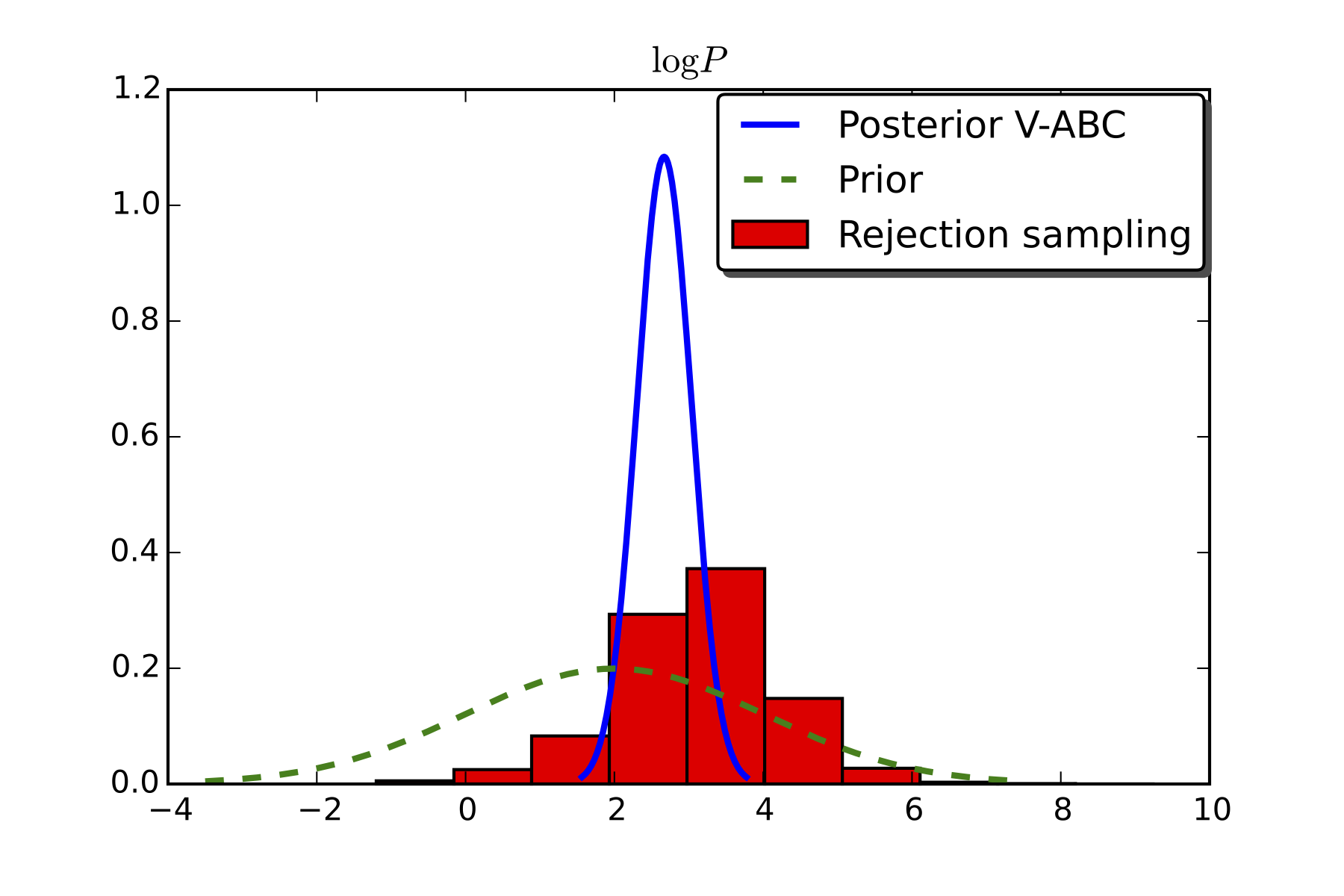}
\label{fig:bf01}}
\quad
\subfigure[$\log \delta$]{%
\includegraphics[width=37mm,height=41mm]{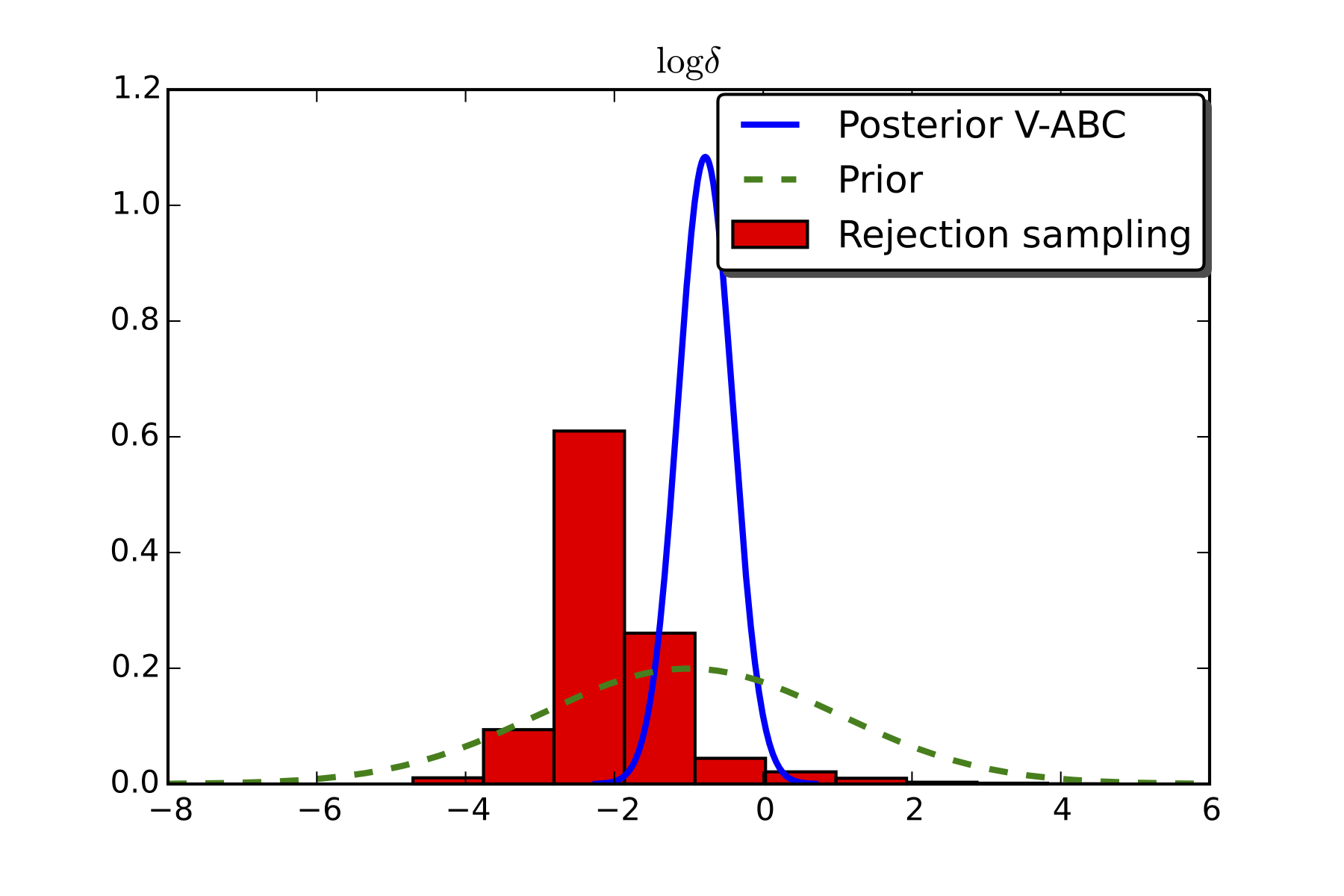}
\label{fig:bf02}}
\subfigure[$\log N_0$]{%
\includegraphics[width=37mm,height=41mm]{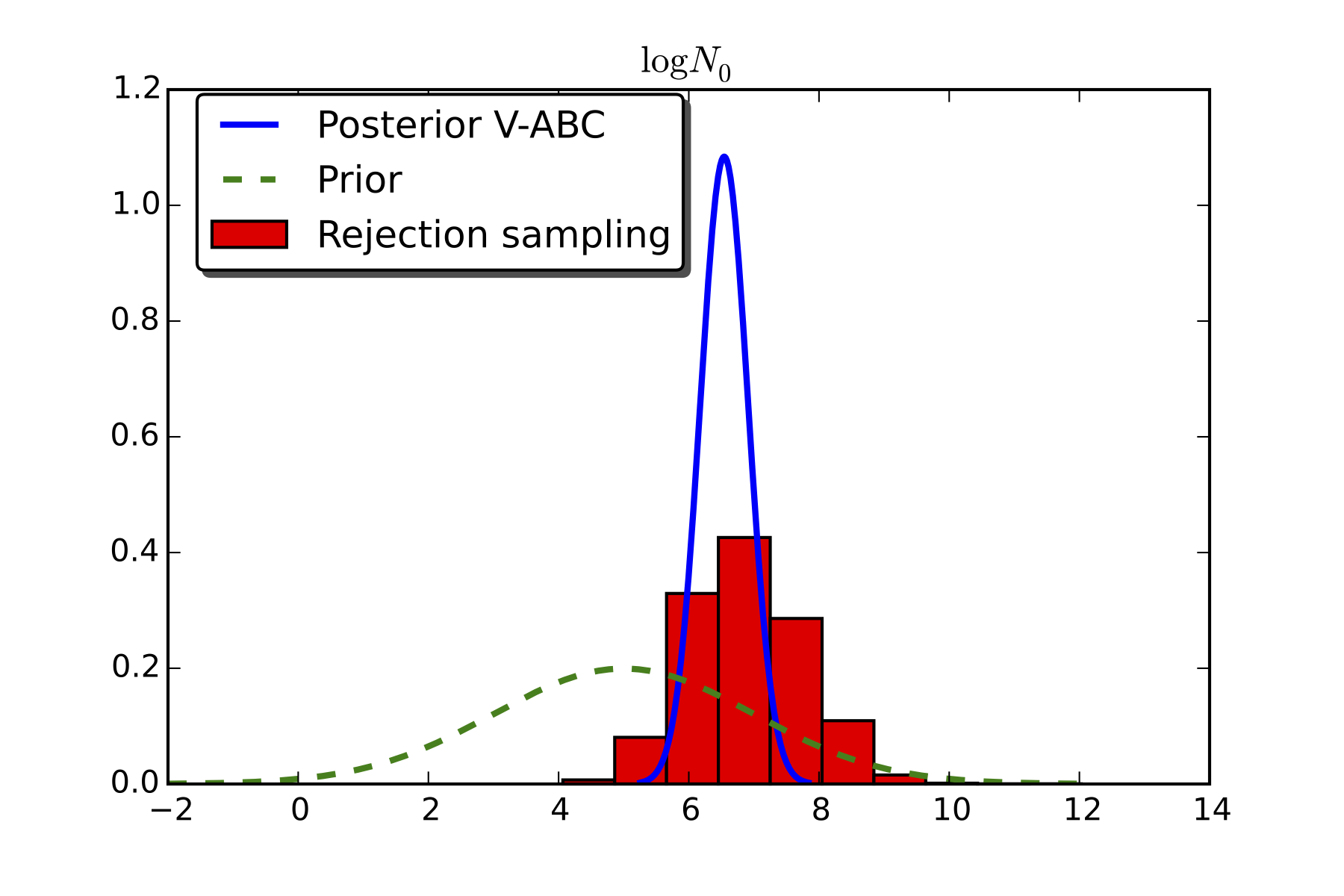}
\label{fig:bf03}}
\quad
\subfigure[$\log \sigma_d$]{%
\includegraphics[width=37mm,height=41mm]{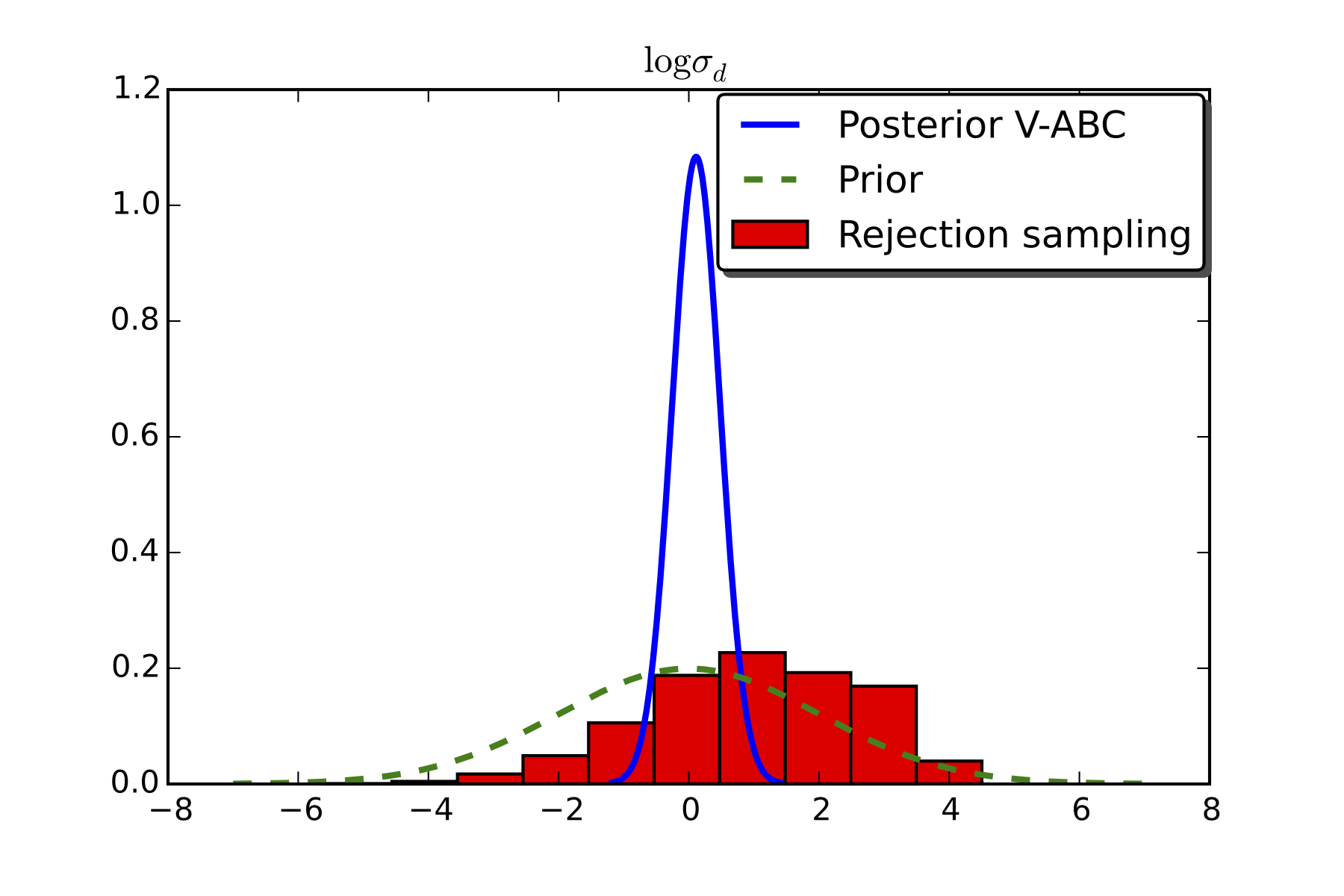}
\label{fig:bf04}}
\quad
\subfigure[$\log \sigma_p$]{%
\includegraphics[width=37mm,height=41mm]{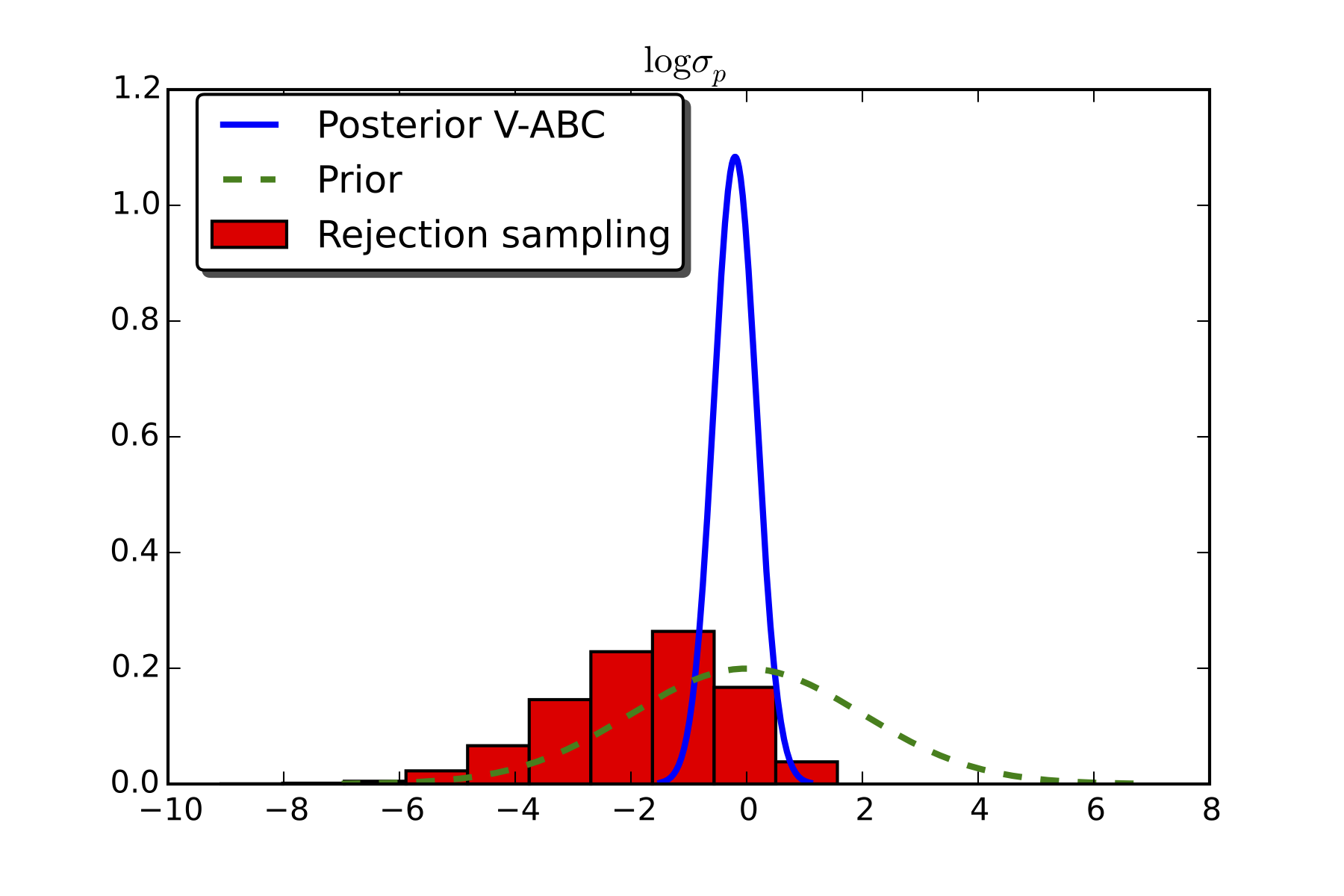}
\label{fig:bf06}}
\quad
\subfigure[Simulation Performance]{\includegraphics[width=40mm,height=41mm]{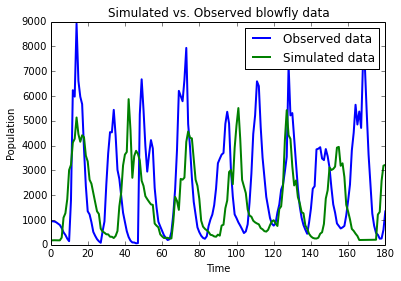}\label{fig:bf07}
\label{fig:bf05}}
\quad
\subfigure[Lower bounds]{
\includegraphics[width=37mm,height=41mm]{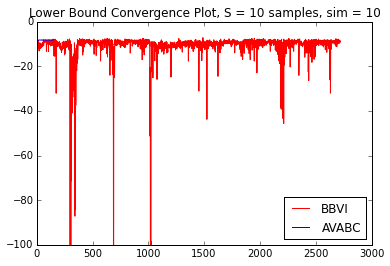}
}
\caption{a-e: Plots for the first 5 parameters of the blowfly experiment: Posterior probability, $p(\theta|\bm{y})$, along with the corresponding prior probabilities and rejection samples produced by $\epsilon=0.25$. For AVABC, 10 simulations per sample are used. f: Simulation performance, plot of simulated vs observed data at each time step g: Lower bounds with 10 samples and 10 simulations per sample.  Variance of AVABC (blue) is considerably lower than BBVI (red). Convergence of AVABC requires fewer iterations than BBVI.
}
\label{fig:bf_thetas}
\end{figure}

For this experiment, we used control variates and ADAM as the only variance reduction for BBVI.  However, both our method and BBVI can be easily extended to use Rao Blackwellization as well.  For the 5 parameters, $\theta_{1 \cdots 5}$, we obtain the posteriors shown in Figures~\ref{fig:bf01}-\ref{fig:bf05}. In Figure~\ref{fig:bf06}, the Blowfly lower bound plot is shown. As can be seen in Figure~\ref{fig:bf06}, the AVABC lower bound has a lower variance and a faster convergence rate, i.e.\ number of required iterations for AVABC before it stabilizes is about $175$, whereas BBVI requires nearly $2750$ iterations in order to stabilize to some extent. In fact as per the current set of experiments performed on blowfly, BBVI did not really reach a high level of stability.

\section{Discussion and Future Work}
In this paper, we introduced Automatic Variational ABC, a low-variance variational likelihood-free inference algorithm.  With our approach, all simulation code, prior and Q distributions, and the variational lower-bound are written within an autodifferentiation environment.  Taking this route towards Bayesian inference of simulation models requires a initial overhead of rewriting simulation code, but reaps the benefits of allowing end-to-end optimization of the lower-bound in an automatic fashion using stochastic gradient ascent.

Our algorithm also uses several reparameterizations of both the simulation--making it a deterministic function of parameters and random latents--and the Q distributions.  These reparameterizations not only allow one to perform automatic inference, but also greatly reduce the variance of the gradients of the lower-bound.  This variance reduction makes variational inference for likelihood-free models feasible.  Further, by using adaptive gradient-step algorithms, such as Adagrad or Adam, learning rates can automatically be  adjusted to the variance of the gradients, which helps significantly for both our algorithm and alternatives.  For variational inference of simulation models, variance reduction is the name of the game.  


We demonstrated performance on three experiments and compared to alternatives BBVI and VBIL; we found that AVABC achieves much lower variance and faster convergence for both toy and real problems.  
Despite the positive preliminary results AVABC demonstrated, our algorithm in the current form has clear limitations.  First, it assumes that one can write the variational posterior distributions as a deterministic function of its parameters and a set of parameterless variables.   Some distributions, such as the Beta or Gamma distribution, cannot be written this way since they involve rejection steps (this could be overcome but it requires further thought).  Alternative distributions such as the Kumaraswamy for the Beta and the Weibull or log-normal for a Gamma must be used.  
%
%
Second, because we must differentiate the simulator, the outputs must be differentiable, which is a restrictive assumption.  Often, we can approximate discrete outputs by a continuous distribution, as we did when approximating the binomial distribution with a normal distribution, but this may not always be the case.  Most significantly, it requires writing possibly very complex simulation code within an auto-dif environment, which may only be possible on a limited set of simulations.   

In the future we would like to expand upon the allowable Q distributions by writing the random number generators within an auto-dif environment.  More challenging simulators, such as state-space models should also be ported.  
 Along the lines of \cite{Meeds2014GpsUai}\cite{wilkinson:2014}, it would be possible add surrogate functions of the simulator and thus call the surrogate instead of the simulator when it has high confidence.  Another, as suggested in O-BBVI \cite{ruiz2016overdispersed}, is to combine importance sampling with the reparametrization of the variational distribution for further variance reduction.



\subsection*{Acknowledgments}

We thank Facebook and Google for their support.  This work was supported in part by the National Institute of Health grant number R01EY013178, the Amsterdam Academic Alliance Data Science (AAA-DS) Program Award to the UvA and VU Universities, the Georgia Tech Executive Vice President of Research Office and the Center for Computational Health, grants from
the Gordon and Betty Moore Foundation and the National Science Foundation and contributions from OCC members
like the University of Chicago.  Finally, we thank the authors of VBIL for providing their code, and Christian A. Naesseth for helpful discussions that led us to realize the relationship between our method and VBIL.

\small
\bibliographystyle{plain}
\bibliography{nips-version}

\section{Relationship between AVABC and VBIL}

In VBIL, they show that minimizing an augmented parameter KL divergence is equivalent to minimizing the KL divergence between the variational distribution and the true posterior.  The KL divergence they minimize is given by
\begin{align}
KL(\lambda)&=\int q_{\lambda}(\theta)g_N (z|\theta)\log \frac{q_\lambda (\theta)}{p(\theta)p_N (y|\theta,z)}dzd\theta
\end{align}

where $q_{\lambda}(\theta)$ is the variational distribution in the original parameter space, $p_N (y|\theta,z)$ is an estimator of the likelihood (for example, in the ABC case $p_N (y|\theta,z)=p_\epsilon(y|\theta)$, the ABC likelihood), $z=\log\frac{p_N(y|\theta,z)}{p(y|\theta)}$, and $g_N (z|\theta)$ is the density for $z$.  The VBIL authors note that calculating $p_N(y|\theta,z)$ implicitly generates $z$.

Now note
\begin{align}
KL(\lambda)&=\int q_{\lambda}(\theta)g_N (z|\theta)\log \frac{q_\lambda (\theta)}{p(\theta)p_N (y|\theta,z)}dzd\theta\\
&=\int q_\lambda(\theta)\log\frac{q_\lambda(\theta)}{p(\theta)}d\theta-\int q_{\lambda}(\theta)g_N (z|\theta)\log p_N(y|\theta,z)dzd\theta\\
&=KL(q(\theta)||p(\theta))-\int q_{\lambda}(\theta)g_N (z|\theta)\log p_N(y|\theta,z)dzd\theta\\
&=KL(q(\theta))||p(\theta))-E_{\theta\sim q(\theta),z\sim g_N(z|\theta)}\left(\log p_N(y|\theta,z)\right)\\
&=-(E_{\theta\sim q(\theta),z\sim g_N(z|\theta)}\left(\log p_N(y|\theta,z)\right)-KL(q(\theta))||p(\theta)))
\end{align}

Note the similarity to equation ~\ref{eq:avabc} in the main paper.  The only differences are that they include the density for $z$, and $p_N(y|\theta,z)$ is any estimator of the likelihood.  Since calculating $p_N(y|\theta,z)$ implicitly generates $z$, our estimator for $\mathcal{L_{ABC}}$ is an unbiased estimator of $-KL(\lambda)$.  Thus the theory they developed in VBIL holds for our problem and the biased gradients due to taking the log of an estimator are not an issue.  The bias introduced by the choice of $\epsilon$ can in theory be an issue, but we find that in practice the choice we described in the main paper gives good results.

\section{Similarity between VBIL and BBVI}

In VBIL, we have $\hat{h}(\theta,z)=\log(p(\theta)\hat{p}_{N}(y|\theta,z))$, where $\hat{p}_{N}(y|\theta,z)$ is an estimator of the likelihood as described above.  Let $q_{\lambda}(\theta)$ be the variational distribution in the original parameter space.  Then, an estimator of the KL divergence between $q_{\lambda,z}(\theta,z)$ and $\pi_{N}(\theta,z)$ in the augmented parameter space is 

\begin{align}
\nabla_{\lambda}KL(\lambda)^{naive}&=\frac{1}{S}\sum_{s=1}^{S}\nabla_{\lambda}[\log q_{\lambda}(\theta^{(s)})](\log q_{\lambda}(\theta^{(s)})-\hat{h}(\theta^{(s)},z^{(s)}))
\end{align}

Note that $z$ is never dealt with explicitly.  It is only handled implicitly via the unbiased estimator to the likelihood.  If the unbiased estimator equals the true likelihood, we get $\hat{h}(\theta,z)=\log(p(\theta)p(y|\theta))=\log(p(y,\theta))$.  We then have

\begin{align}
\nabla_{\lambda_{i}}KL(\lambda)^{naive}&=\frac{1}{S}\sum_{s=1}^{S}\nabla_{\lambda}[\log q_{\lambda}(\theta^{(s)})](\log q_{\lambda}(\theta^{(s)})-\log p(y,\theta^{(s)}))\\
\end{align}

The estimator for BBVI is

\begin{align}
\nabla_{\lambda}\mathcal{L}\approx \frac{1}{S}\sum_{s=1}^{S}\nabla_{\lambda}\log q_{\lambda}(\theta^{(s)})(\log p(y,\theta^{(s)})-\log q_{\lambda}(\theta^{(s)}))
\end{align}

Note that the VBIL estimator under the true likelihood is the negative of the BBVI estimator.  Building from this, VBIL has one major difference: they use natural gradient descent \cite{amari1998natural} instead of ADAGRAD \cite{duchi2011adaptive} or ADAM \cite{kingma2014adam}.

\section{Learning Latent Variables per Datapoint}
We start with the marginal probability as before, but introduce latent variables $z$ (no relationship with the $z$ described for VBIL) and assume a factorized variational posterior $Q(\theta,z|\phi,\xi)=Q_\phi(\theta)\prod_n Q_\xi(z_n)$ and factorized prior $p(\theta,z)=p(\theta)p(z)$.
\begin{align}
\log p(y)&=\log \int p(y|\theta,z)p(\theta)p(z)d\theta dz\\
&=\log \int Q_\phi(\theta)Q_\xi(z)\frac{p(y|\theta,z)p(\theta)p(z)}{Q_\phi(\theta)Q_\xi(z)}d\theta dz\\
&=\log \mathbb{E}_{Q_\phi(\theta),Q_\xi(z)}\left( \frac{p(y|\theta,z)p(\theta)p(z)}{Q_\phi(\theta)Q_\xi(z)}\right)\\
&\geq \mathbb{E}\left(\log \frac{p(y|\theta,z)p(\theta)p(z)}{Q_\phi(\theta)Q_\xi(z)}\right)\\
&=\mathbb{E}(\log p(y|\theta,z))-KL(Q_\phi(\theta)||p(\theta))- KL(Q_\xi(z)||p(z))\\
\end{align}
We again apply the ABC likelihood and the reparametrization to make the simulator deterministic
\begin{align}
\mathcal{L_{ABC}}&=\mathbb{E}_{Q_\phi(\theta)Q_\xi(z)}(\log p_\epsilon(y|\theta,z))-KL(Q_\phi(\theta)||p(\theta))- KL(Q_\xi(z)||p(z))\\
&=\mathbb{E}(\log \int p_\epsilon(y|f(\theta,z,u))p(u)du)-KL(Q_\phi(\theta)||p(\theta))- KL(Q_\xi(z)||p(z))\\
&=\mathbb{E}(\log \int p_{e}(y|f(g(\phi,\nu),h(\xi,w),u)p(u)du)-KL(Q_\phi(\theta)||p(\theta))- KL(Q_\xi(z)||p(z))
\end{align}
We can again replace all expectations with samples to obtain
\begin{align}
\mathcal{L_{ABC}}&\approx\frac{1}{S}\frac{1}{K}\sum_s\sum_k\log \frac{1}{L}\sum_l p_\epsilon(y|f(g(\phi,\nu^{(s)}),h(\xi,w^{(k)}),u^{(l)})-KL(Q_\phi(\theta)||p(\theta))\nonumber\\
&\qquad- KL(Q_\xi(z)||p(z))
\end{align}

\end{document}